\documentclass[letterpaper, 10 pt, conference]{ieeeconf}  
\IEEEoverridecommandlockouts          
\usepackage{amsmath,amsfonts}
\usepackage{algorithmic}
\usepackage{array}
\usepackage[caption=false,font=normalsize,labelfont=sf,textfont=sf]{subfig}
\usepackage{textcomp}
\usepackage{marvosym}
\usepackage{stfloats}
\usepackage{url}
\usepackage{verbatim}
\usepackage{graphicx}
\usepackage{multirow}
\usepackage{booktabs}
\usepackage{caption}
\usepackage{lipsum}
\usepackage{balance}
\usepackage{hyperref}
\usepackage{CJKutf8}
\usepackage{cite}
\usepackage{url}
\hypersetup{
	colorlinks=true,           
	linkcolor=red,              
	citecolor=green,           
	filecolor=blue,  
	urlcolor=blue           
}

\usepackage{url}
\usepackage{soul}

\usepackage{graphicx} 
\usepackage{amssymb}
\usepackage{makecell}

\usepackage{wrapfig}
\usepackage{booktabs}
\usepackage{tikz}
\usepackage{circledsteps}
\let\labelindent\relax
\usepackage{enumitem}

\usepackage{amsmath,amssymb,mathtools}

\usepackage{bm}
\usepackage{amsmath}
\usepackage{algorithm}
\usepackage{multicol,multirow}
\usepackage{hyperref}
\usepackage[table]{xcolor}

\renewcommand{\footnoterule}{%
  \kern-3pt
  \hrule width 0.4\columnwidth
  \kern 2.6pt
}

\title{\LARGE \bf
Diagnosing and Dynamically Filtering Occupancy World Models \\ for Active Mapping}

\author{Jiahui Zhang$^{1}$, Gongbo Liang$^{2}$, and Yu Zhang$^{1 \text{\Letter}}$%
\thanks{This work is supported by NSF Grant No. 2429968.}%
\thanks{$^{1}$Boise State University}%
\thanks{$^{2}$Texas A\&M University--San Antonio}%
\thanks{$^{\text{\Letter}}$Correspondence to: Yu Zhang: \texttt{yzhang@boisestate.edu}}}

\begin{document}

\maketitle
\thispagestyle{empty}
\pagestyle{empty}


\begin{abstract}
Active mapping requires a robot to select camera viewpoints that efficiently reconstruct an unknown 3D scene. To reason about unobserved regions, recent systems use pretrained occupancy networks as world models that complete missing geometry. The predicted structure contributes to expected coverage gain and constrains feasible robot motion. Consequently, occupancy errors can change both what the robot chooses to explore and where it is able to move. We diagnose these effects by holding the planner fixed and varying only the occupancy representation provided to it. We consider planning without completion, with learned occupancy, with false positives removed by a ground truth oracle, with false negatives restored by an oracle, and with ground truth occupancy. Our experiments show that correcting false positives or false negatives alone does not consistently improve final coverage. This finding reveals a gap between occupancy accuracy and downstream planning performance. Ground truth occupancy provides a much larger improvement in coverage efficiency than in endpoint coverage, suggesting that planning and reachability remain important bottlenecks even when the geometric world model is accurate. Based on these findings, we introduce a dynamic filtering strategy that preserves predictions in unexplored space while suppressing repeatedly unsupported occupancy using online observations. Preliminary examples show that this strategy can redirect viewpoint selection toward reachable surfaces that would otherwise remain unobserved.

\end{abstract}

\begin{figure}[t!]
    \centering
    \includegraphics[width=\linewidth]{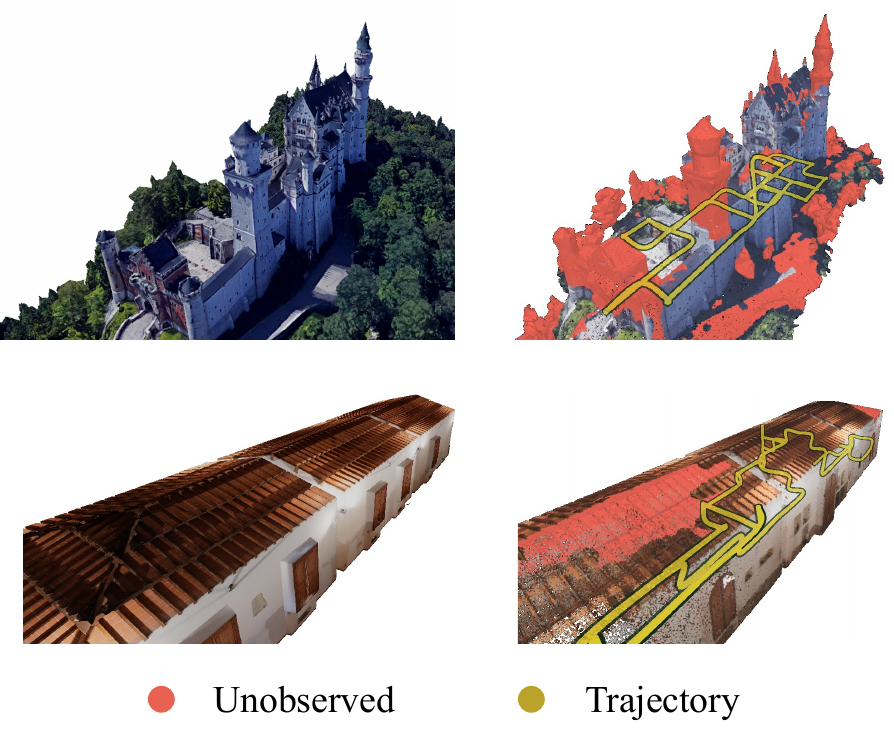}
    \caption{\textbf{Active mapping and missed surface coverage.} The left column shows the complete reference scenes, and the right column shows the coverage produced by representative active mapping trajectories. Yellow curves indicate the camera trajectories. Surfaces retaining their original colors were observed, while red surfaces remained unobserved. These examples illustrate how viewpoint selection under a limited motion budget can leave scene geometry uncovered.}
    \label{fig:teaser}

\vspace{-0.3cm}
\end{figure}
\section{Introduction}

Active mapping~\cite{lluvia2021active} asks a robot to reconstruct an unknown 3D scene while deciding where to observe next. Because future measurements depend on the selected camera path, inefficient planning can leave substantial geometry unobserved~\cite{li2026magician}. Figure~\ref{fig:teaser} illustrates this problem. The yellow curves show the camera trajectories, while the red surfaces show geometry missed by those trajectories. The goal is to maximize surface coverage under a limited motion budget.

Planning is difficult because the robot initially observes only part of the scene~\cite{wu2025embodiedocc}~\cite{zhang2023crossadapt}~\cite{zhang2023crossseg}~\cite{zhang2025static}. Recent systems use pretrained occupancy networks to complete unobserved geometry~\cite{li2026magician}~\cite{guedon2023macarons}. The predicted occupancy acts as a geometric world model in two ways. It provides expected coverage gain for candidate viewpoints and constrains collision-free motion. A false positive may create gain around nonexistent geometry or block a free path, while a false negative may hide a useful target or alter apparent reachability.

Occupancy accuracy alone does not reveal these planning effects~\cite{mescheder2019occupancy}. Active mapping is a closed-loop process in which each viewpoint changes later observations, predictions, and decisions~\cite{li2026magician}. A more accurate occupancy map can therefore produce a different trajectory without improving final coverage. The world model must be evaluated through its effect on the complete exploration process.

We study this interaction using MAGICIAN~\cite{li2026magician} on 25 scene-start configurations from the Macarons++ benchmark. We hold the planner and all other components fixed while varying only the occupancy supplied to it, including learned occupancy, no completion, oracle correction of false positives or false negatives, and ground-truth occupancy. We also decompose the planner's rendered gain to measure how phantom predictions enter viewpoint scoring. The results show that the value of completion depends on the starting condition, correcting one error type does not produce a monotonic improvement, and ground-truth occupancy benefits coverage efficiency much more than endpoint coverage.

Motivated by these findings, we introduce an observation-gated dynamic filter. It retains predictions in insufficiently explored regions and suppresses them from both gain estimation and collision checking only after repeated frustum exposure without nearby RGB-D surface support. The filter requires no ground truth or retraining. On two collapse-prone starts, it improves mean final coverage by $0.163$ and $0.122$ over three repetitions.

Our contributions are summarized as follows.

\begin{itemize}

\item We provide a controlled diagnosis of how occupancy completion and its errors affect active mapping, combining decision-level gain decomposition with fixed-planner occupancy interventions.

\item We show that occupancy accuracy and planning performance are not equivalent. The effects depend on the starting condition and differ between trajectory efficiency and final coverage.

\item We introduce an observation-gated dynamic filter that uses online RGB-D observations to suppress repeatedly unsupported occupancy, with preliminary improvements on two targeted failure cases.

\end{itemize}

\section{Related Work}
\subsection{Occupancy Networks}

Neural implicit models encode geometry as continuous occupancy or distance functions. Occupancy Networks introduced continuous inside--outside classification~\cite{mescheder2019occupancy}; IF-Net and Convolutional Occupancy Networks added spatially aligned features for detailed object and scene reconstruction~\cite{chibane2020implicit}. Neural-Pull~\cite{ma2020neural} and POCO~\cite{boulch2022poco} reconstruct implicit surfaces from point clouds, while MonoScene~\cite{cao2022monoscene}, VoxFormer~\cite{li2023voxformer}, and OccFormer~\cite{zhang2023occformer} predict dense semantic occupancy from images.

For embodied tasks, Occupancy Anticipation completes unseen free and occupied space~\cite{ramakrishnan2020occupancy}, whereas NICE-SLAM~\cite{zhu2022nice} and iSDF~\cite{ortiz2022isdf} update implicit geometry online~\cite{zhang2021dynamic}. These works improve geometric representation and prediction. We instead examine occupancy as a planner-facing world model whose errors alter both coverage gain and collision checking, and therefore the complete trajectory.

\subsection{Active Mapping}

Active mapping selects observations under sensing and motion budgets. Classical methods combine information or frontier objectives with path planning. PC-NBV~\cite{zeng2020pc} learns view selection from partial point clouds, while FUEL~\cite{zhou2021fuel} and TARE~\cite{cao2021tare} organize large-scale exploration; uncertainty-guided systems instead target uncertain geometry or scene properties~\cite{georgakis2022uncertainty}~\cite{haughton2022real}.

Neural methods increasingly plan over learned scene representations. ActiveNeRF~\cite{pan2022activenerf} uses radiance-field uncertainty, while SCONE~\cite{guedon2022scone} and MACARONS~\cite{guedon2023macarons} use learned occupancy and coverage prediction for view selection. Active Neural Mapping~\cite{yan2023active}, NARUTO~\cite{feng2024naruto}, and FisherRF~\cite{jiang2024fisherrf} guide motion from map uncertainty or information, and NextBestPath~\cite{li2025nextbestpath} plans beyond a single view. MAGICIAN~\cite{li2026magician} converts pretrained occupancy into imagined Gaussians for long-horizon trajectory search, coupling completion with rendered gain and collision checking. We keep this planner fixed and vary only its occupancy input to isolate the effects of hallucinated geometry and motivate dynamic filtering.

\section{Diagnosing Occupancy World Models}
\label{sec:diagnosis}

\subsection{Problem Formulation}

We formulate the problem by folowing prior active mapping works~\cite{li2026magician}~\cite{guedon2023macarons}. At planning step $t$, the agent has accumulated a surface point cloud $\mathcal{S}_t$ from its RGB-D observations and a set of previously visited camera poses $\mathcal{C}_t$. A pretrained occupancy network predicts the probability that a query point $\mathbf{x}$ is occupied,
\begin{equation}
\hat{\sigma}_t(\mathbf{x})
=
f_{\theta}(\mathbf{x},\mathcal{S}_t,\mathcal{C}_t),
\qquad
\hat{\sigma}_t(\mathbf{x})\in[0,1].
\end{equation}
We denote the thresholded predicted occupancy by
\begin{equation}
\widehat{\mathcal{O}}_t
=
\left\{\mathbf{x}\mid\hat{\sigma}_t(\mathbf{x})>0.5\right\}.
\end{equation}
These predictions complete geometry beyond the currently observed surface and serve as a world model for planning.

We instantiate our diagnosis using MAGICIAN~\cite{li2026magician}, which consumes the predicted occupancy through two coupled channels. First, predicted occupied points are represented as Gaussian primitives and rendered from each candidate viewpoint. The rendered novelty map produces the predicted coverage gain
\begin{equation}
G_t(\mathbf{c};\widetilde{\mathcal{O}}_t)
=
\sum_{\mathbf{p}\in\mathcal{P}_{\mathrm{valid}}}
w\!\left(D_{\mathbf{c}}(\mathbf{p})\right)
I^{\mathrm{novel}}_t
\!\left(\mathbf{p};\widetilde{\mathcal{O}}_t\right),
\label{eq:gain}
\end{equation}
where $\widetilde{\mathcal{O}}_t$ is the occupancy supplied to the planner, $I^{\mathrm{novel}}_t$ is the rendered contribution of predicted surfaces that have not been covered, and $w(\cdot)$ is the planner's depth-dependent weighting. Second, the same occupancy is used during collision checking. The planner therefore selects a trajectory schematically as
\begin{equation}
\boldsymbol{\tau}^{*}_t
=
\arg\max_{\boldsymbol{\tau}\in\mathcal{T}_t}
\sum_{\mathbf{c}\in\boldsymbol{\tau}}
G_t(\mathbf{c};\widetilde{\mathcal{O}}_t)
\quad
\mathrm{s.t.}\quad
\operatorname{Free}
(\boldsymbol{\tau};\widetilde{\mathcal{O}}_t)=1.
\label{eq:planning}
\end{equation}
Consequently, an occupancy prediction can affect both the score of a candidate trajectory and whether that trajectory is considered feasible. A false positive can generate gain around nonexistent geometry or block a free path, while a false negative can remove a useful target or alter apparent reachability. The net effect must therefore be measured through the resulting closed-loop trajectory rather than occupancy accuracy alone.

\subsection{Dataset}

\begin{figure*}[!t]
\centering
\includegraphics[width=\textwidth]{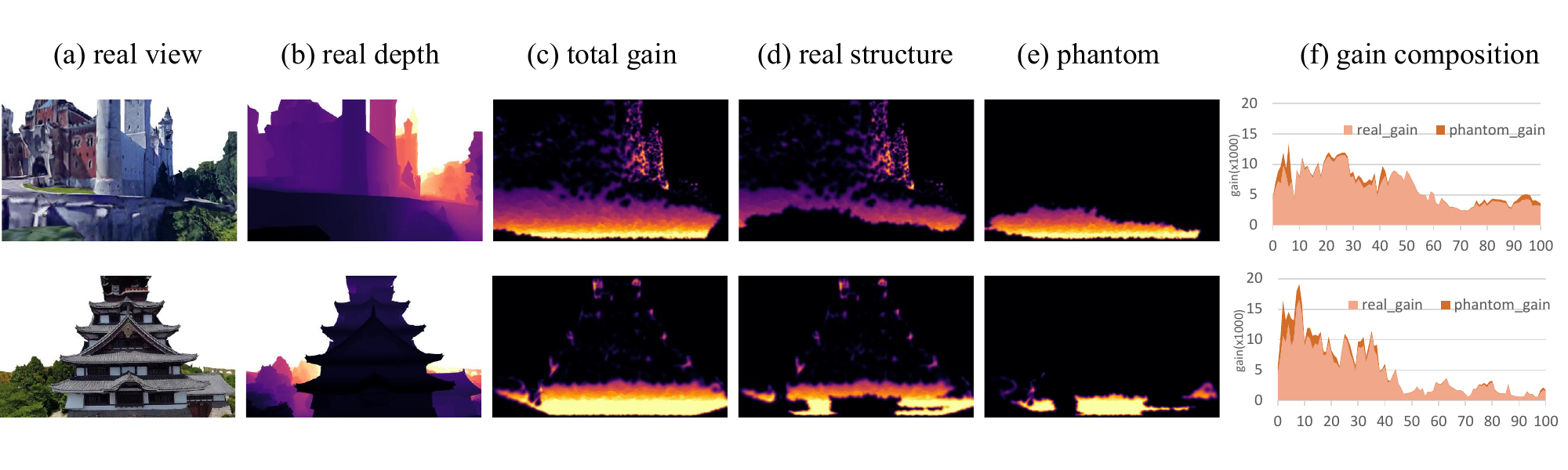}
\caption{\textbf{Decision-level audit of predicted coverage gain.} Top: Neuschwanstein Castle at start 3. Bottom: Fushimi Momoyama Castle at start 0. Each row analyzes a viewpoint selected by the learned-occupancy planner at steps 6 and 2, respectively. Panels (a) and (b) show the RGB and depth observations from that pose, while (c) shows the per-pixel coverage gain used by the planner. For analysis only, ground-truth tags decompose the same rendered gain into (d) predictions near real structure and (e) phantom predictions, with (c) equal to (d) plus (e) pixel-wise on a shared color scale. Phantom occupancy contributes 53.5\% and 35.0\% of the gain in the displayed frames respectively. Panel (f) shows the same decomposition over all steps.}
\label{fig:audit}
\end{figure*}

We conduct all experiments on Macarons++~\cite{li2026magician}, an extension of the Macarons benchmark~\cite{guedon2023macarons} containing large-scale indoor and outdoor 3D scenes. The benchmark supports free six-degree-of-freedom exploration and provides RGB-D observations from arbitrary camera poses. We use five scenes with five fixed starting poses per scene, resulting in 25 scene-start configurations. Each trajectory contains 100 planning actions and 101 camera poses. The planner receives only the RGB-D observations generated along its trajectory. Direct access to the ground-truth mesh is restricted to coverage evaluation, the diagnostic visualizations in Figure~\ref{fig:audit}, and the oracle interventions in Table~\ref{tab:interventions}.

\subsection{Controlled Occupancy Interventions}

To isolate this effect, we hold the occupancy network, beam search, candidate viewpoints, collision rule, motion budget, and reconstruction pipeline fixed. We vary only $\widetilde{\mathcal{O}}_t$, the occupancy supplied to the planner at each step.

Let $\mathcal{O}^{*}$ denote the ground-truth occupied samples under the benchmark discretization. For diagnostic analysis, a predicted point is classified as a false positive when it lies farther than $\delta=0.03d_{\mathrm{scene}}$ from the ground-truth surface. A ground-truth point is classified as a false negative when no predicted point lies within the same tolerance:
\begin{align}
\mathrm{FP}_t
&=
\left\{
\mathbf{x}\in\widehat{\mathcal{O}}_t
\mid
d(\mathbf{x},\mathcal{O}^{*})>\delta
\right\},\\
\mathrm{FN}_t
&=
\left\{
\mathbf{y}\in\mathcal{O}^{*}
\mid
d(\mathbf{y},\widehat{\mathcal{O}}_t)>\delta
\right\}.
\end{align}
These labels require ground truth and are used only to construct diagnostic oracles. We compare the following five planning conditions:
\begin{align}
\widetilde{\mathcal{O}}^{\mathrm{no\mbox{-}imag}}_t
&=
\mathcal{S}_t,\\
\widetilde{\mathcal{O}}^{-\mathrm{FP}}_t
&=
\mathcal{S}_t
\cup
\left(
\widehat{\mathcal{O}}_t\setminus\mathrm{FP}_t
\right),\\
\widetilde{\mathcal{O}}^{\mathrm{base}}_t
&=
\mathcal{S}_t\cup\widehat{\mathcal{O}}_t,\\
\widetilde{\mathcal{O}}^{+\mathrm{FN}}_t
&=
\mathcal{S}_t
\cup
\widehat{\mathcal{O}}_t
\cup
\mathrm{FN}_t,\\
\widetilde{\mathcal{O}}^{\mathrm{GT}}_t
&=
\mathcal{O}^{*}.
\end{align}
The no-imagination condition retains only accumulated observations. The $-\mathrm{FP}$ condition removes predicted geometry unsupported by ground truth. The $+\mathrm{FN}$ condition restores missed ground-truth geometry while retaining the model's false positives. The final condition gives the fixed planner complete ground-truth occupancy. Each condition is executed closed-loop, so different planning decisions lead to different subsequent observations and world-model updates.

We evaluate 25 starts from five scenes, with 101 camera poses per trajectory. Let $C_t$ denote surface coverage after step $t$. We report final coverage $C_T$, the normalized area under the coverage trajectory,
\begin{equation}
\mathrm{AUC}
=
\frac{1}{T}\int_{0}^{T}C(t),dt,
\end{equation}
and the first step at which a method reaches $70\%$ of the learned baseline's final coverage at the same start,
\begin{equation}
S_{70}
=
\min\left\{
t\mid C_t\geq0.7C_T^{\mathrm{base}}
\right\}.
\end{equation}
Final coverage measures the endpoint, while AUC and $S_{70}$ measure how efficiently coverage is acquired. All reported differences are computed relative to the learned-occupancy baseline at the same start before averaging.

For descriptive analysis, we additionally separate starts where the baseline ends below $0.75$ coverage and above $0.85$ coverage. We refer to these as baseline low-coverage and high-coverage starts. This grouping is defined from the baseline outcome and is used to expose heterogeneity, not to claim that failure can be predicted from the initial pose.

\begin{table*}[t]
\centering
\small
\setlength{\tabcolsep}{4pt}
\caption{\textbf{Downstream effects of controlled occupancy interventions.} We hold the planner fixed and vary only the occupancy supplied to it. The learned-occupancy row reports absolute values, while all other rows report mean paired differences relative to that baseline. Results are shown for all starts ($n{=}25$), baseline low-coverage starts ($C_T^{\mathrm{base}}<0.75$, $n{=}5$), and baseline high-coverage starts ($C_T^{\mathrm{base}}>0.85$, $n{=}16$). The four intermediate starts are included only in the pooled results. $S_{70}$ is the number of steps required to reach 70\% of the baseline final coverage at the same start. Lower values and negative differences in $S_{70}$ indicate faster coverage. The FP, FN, and ground-truth conditions use ground truth only as diagnostic oracles.}
\label{tab:interventions}
\begin{tabular}{@{}l c ccc ccc ccc@{}}
\toprule
& &
\multicolumn{3}{c}{Pooled ($n{=}25$)} &
\multicolumn{3}{c}{Low coverage ($n{=}5$)} &
\multicolumn{3}{c}{High coverage ($n{=}16$)}\\
\cmidrule(lr){3-5}
\cmidrule(lr){6-8}
\cmidrule(lr){9-11}
Planning occupancy &
Oracle &
$\Delta C_T\,\uparrow$ & $\Delta\mathrm{AUC}\,\uparrow$ & $\Delta S_{70}\,\downarrow$ &
$\Delta C_T\,\uparrow$ & $\Delta\mathrm{AUC}\,\uparrow$ & $\Delta S_{70}\,\downarrow$ &
$\Delta C_T\,\uparrow$ & $\Delta\mathrm{AUC}\,\uparrow$ & $\Delta S_{70}\,\downarrow$\\
\midrule
(i) no imagination
& no
& $-0.007$ & $-0.038$ & $+4.6$
& $+0.100$ & $+0.055$ & $-8.6$
& $-0.050$ & $-0.073$ & $+9.4$\\

(ii) occ $-$ FP
& yes
& $-0.004$ & $-0.006$ & $+2.5$
& $+0.020$ & $+0.019$ & $-0.6$
& $-0.029$ & $-0.019$ & $+4.2$\\

\midrule
(iii) learned occ. (base, absolute)
& no
& $0.846$ & $0.644$ & $31.6$
& $0.676$ & $0.485$ & $39.6$
& $0.908$ & $0.706$ & $27.4$\\
\midrule

(iv) occ $+$ FN
& yes
& $-0.026$ & $+0.031$ & $-3.0$
& $-0.044$ & $+0.035$ & $-4.0$
& $-0.023$ & $+0.030$ & $-1.9$\\

(v) ground truth
& yes
& $+0.031$ & $+0.086$ & $-12.7$
& $+0.095$ & $+0.178$ & $-28.4$
& $+0.004$ & $+0.055$ & $-7.4$\\
\bottomrule
\end{tabular}
\end{table*}

\subsection{Diagnostic Findings}

\paragraph{Finding 1. Phantom occupancy enters the planner's decision signal.} Figure~\ref{fig:audit} decomposes the exact coverage gain used for viewpoint selection. In the two representative selected views, phantom occupancy contributes 53.5\% and 35.0\% of the predicted gain, so the phantom component is substantial even when the planner predicts high information value. Panel (f) further shows that phantom gain is present across the planning horizon. The figure establishes how occupancy errors enter viewpoint scoring, while Table~\ref{tab:interventions} measures their effect on the resulting trajectories.

\paragraph{Finding 2. The value of completion depends on the starting condition.} Removing completion changes pooled final coverage by only $-0.007$, but this average hides opposing effects. On baseline low-coverage starts, it improves final coverage by $0.100$, increases AUC by $0.055$, and reaches the coverage milestone $8.6$ steps earlier. On high-coverage starts, it reduces final coverage by $0.050$, decreases AUC by $0.073$, and requires $9.4$ additional steps. Completion therefore helps some trajectories explore efficiently while degrading others. Because these groups are defined from baseline outcomes, the result demonstrates heterogeneous behavior rather than an ability to predict failure from the initial pose.

\paragraph{Finding 3. Better occupancy does not translate monotonically into better active mapping.} Oracle removal of false positives does not improve pooled performance, changing final coverage by $-0.004$, AUC by $-0.006$, and $S_{70}$ by $+2.5$ steps. Restoring false negatives produces a different tradeoff. It improves pooled AUC by $0.031$ and reaches the coverage milestone $3.0$ steps earlier, but reduces final coverage by $0.026$. Even ground-truth occupancy benefits efficiency much more than the endpoint. It improves pooled AUC by $0.086$ and reaches the milestone $12.7$ steps earlier, while final coverage increases by only $0.031$. On baseline low-coverage starts, ground-truth occupancy reduces $S_{70}$ from $39.6$ to $11.2$ steps. These results show that occupancy corrections can affect exploration speed and final coverage in different directions. Ground-truth occupancy is therefore a diagnostic reference for the fixed planner rather than a theoretical upper bound on coverage.

\paragraph{Implication.} The results support neither always trusting completion nor discarding it entirely. A practical correction should preserve predictions that guide exploration in unobserved regions and suppress them only after online observations repeatedly fail to provide support. This motivates the dynamic filtering strategy introduced next.

\section{Observation-Gated Dynamic Filtering}
\label{sec:dynamic_filter}

Our diagnosis shows that removing occupancy completion globally can discard useful exploration guidance. We instead filter predictions selectively using two signals available during mapping. For each predicted occupied point, we count how many acquired camera frustums have contained it and check whether a nearby surface has appeared in the accumulated RGB-D observations. A point is marked as unsupported only when it has entered at least $K$ frustums and no observed surface lies within distance $\epsilon$. We use $K=2$ and $\epsilon=0.03d_{\mathrm{scene}}$. Predictions with fewer than $K$ exposures remain unchanged, allowing them to continue guiding exploration in insufficiently observed regions.

Unsupported points are removed from both uses of the occupancy world model. They no longer contribute to rendered coverage gain and are excluded from collision checking. The filter is recomputed after every observation, so a point is restored if later RGB-D measurements provide nearby surface support. The procedure requires no ground truth, model retraining, or additional rendering pass.

\begin{table}[t]
\centering
\small
\setlength{\tabcolsep}{5pt}
\caption{\textbf{Targeted evaluation on two collapse-prone starts.} Final coverage is reported as mean$\pm$std over three repetitions. MAGICIAN~\cite{li2026magician} is used as the baseline.}
\label{tab:dynamic_filter}
\begin{tabular}{@{}lccc@{}}
\toprule
Start & Baseline & Dynamic filter & $\Delta C_T$\\
\midrule
Pantheon/2
& $0.309\pm0.021$
& $\mathbf{0.472\pm0.078}$
& $\mathbf{+0.163}$\\
Sestino/2
& $0.836\pm0.108$
& $\mathbf{0.958\pm0.021}$
& $\mathbf{+0.122}$\\
\bottomrule
\end{tabular}
\end{table}

Table~\ref{tab:dynamic_filter} provides a focused evaluation on the two collapse-prone starts identified by the diagnostic study. Dynamic filtering increases final coverage by $0.163$ on Pantheon/2 and $0.122$ on Sestino/2, providing preliminary evidence that removing repeatedly unsupported occupancy can recover coverage in the targeted failure cases.

\section{Conclusion}

This work diagnoses how pretrained occupancy affects active mapping once used for planning. Phantom geometry changes viewpoint scores, completion can help or harm across starts, and correcting individual error types does not guarantee better trajectories. Ground-truth occupancy mainly improves efficiency, showing that map accuracy is only one planning factor. Our observation-gated filter removes repeatedly unsupported occupancy from gain and collision reasoning, improving both tested collapse-prone starts without retraining or ground truth. These results motivate decision-level evaluation and observation-aware correction of world models as new evidence accumulates.






\newpage
\bibliographystyle{ieeetr} 
\bibliography{main}

\end{document}